\pdfoutput=1
\documentclass[11pt,a4paper]{article}
\usepackage[hyperref]{naaclhlt2018}
\usepackage{times}
\usepackage{latexsym}
\usepackage{scalerel}
\usepackage{url}
\usepackage{amsmath}
\usepackage{amssymb}
\usepackage{subcaption}
\usepackage{float}
\usepackage{color}
\usepackage{multirow}
\usepackage{comment}
\usepackage{algorithm}
\usepackage[noend]{algpseudocode}

\newcommand{\mysubsection}[1]{\vspace{0.3em} \noindent\textbf{#1}}

\aclfinalcopy 

\title{Speaker Naming in Movies}

\author{Mahmoud Azab, Mingzhe Wang, Max Smith, Noriyuki Kojima, Jia Deng, Rada Mihalcea \\
Computer Science and Engineering, University of Michigan\\
{\tt \{mazab,mzwang,mxsmith,kojimano,jiadeng,mihalcea\}@umich.edu}\\
}

\date{}

\begin{document}
\maketitle

\begin{abstract}
We propose a new model for speaker naming in movies that leverages visual, textual, and acoustic modalities in an unified optimization framework. To evaluate the performance of our model, we introduce a new dataset  consisting of six episodes of the Big Bang Theory TV show and eighteen full movies covering different genres. Our experiments show that our multimodal model significantly outperforms several competitive baselines on the average weighted F-score metric. To demonstrate the effectiveness of our framework, we design an end-to-end memory network model that leverages our speaker naming model and achieves state-of-the-art results on the subtitles task of the MovieQA 2017 Challenge.

\end{abstract}

\section{Introduction}

Identifying speakers and their names in movies, and videos in general, is a primary task for many video analysis problems, including automatic subtitle labeling \cite{Hu15}, content-based video indexing and retrieval \cite{Zhang09}, video summarization \cite{Tapaswi14}, and video storyline understanding \cite{Tapaswi14}. It is a very challenging task, as the visual appearance of the characters changes over the course of the movie due to several factors such as scale, clothing, illumination, and so forth \cite{Arandjelovic05,Everingham06}. The annotation of movie data with speakers' names can be helpful in a number of applications, such as movie question answering \cite{Tapaswi16}, automatic identification of character relationships \cite{Zhang09}, or automatic movie captioning \cite{Hu15}.

\begin{figure}[t]
\begin{center}
\scalebox{0.99}{
   \includegraphics[width=\linewidth]{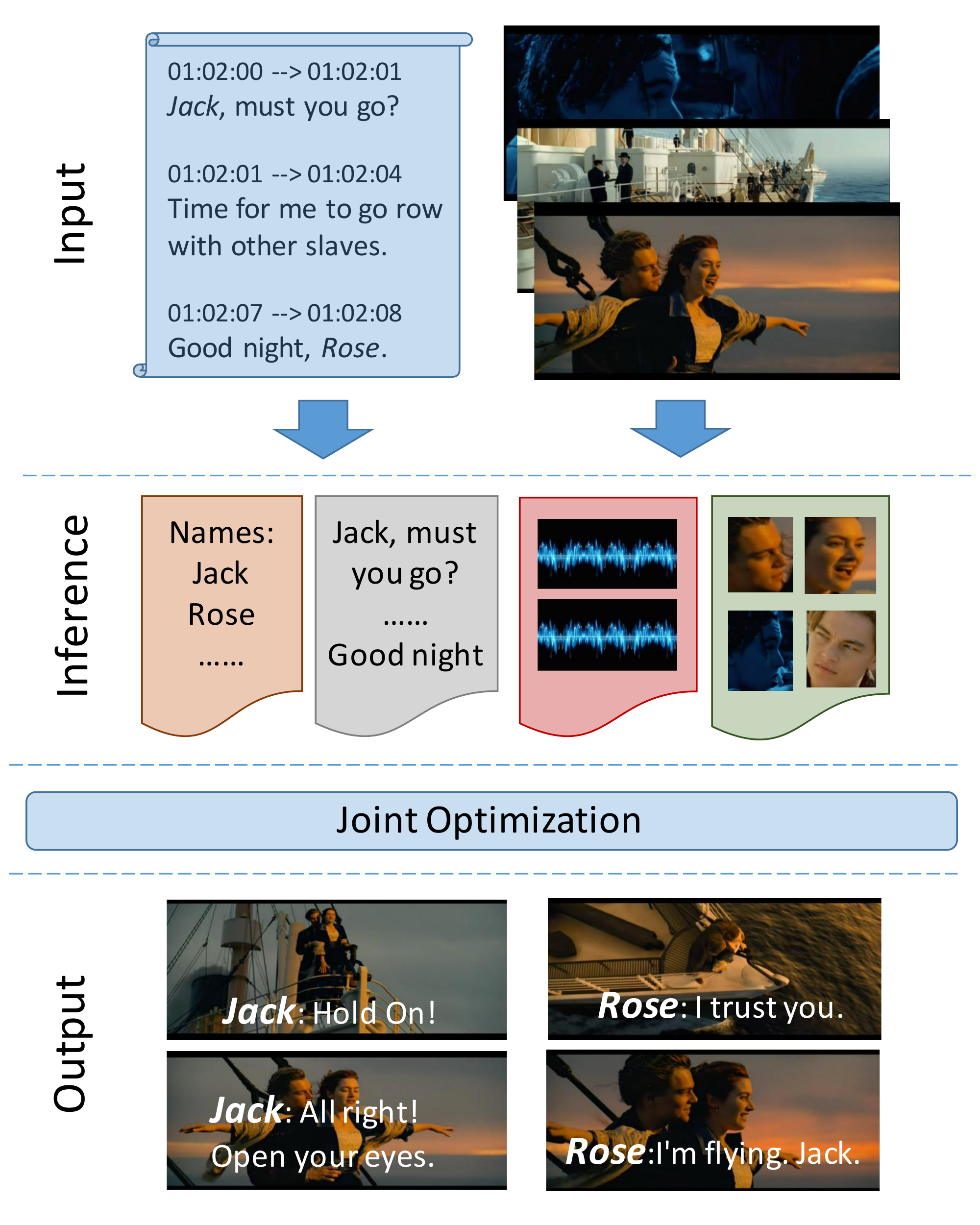}
   }
\end{center}
   \vspace{-0.2cm}
   \caption{Overview of our approach for speaker naming.}
   \vspace{-0.2cm}
\label{fig:long}
\end{figure}

Most previous studies relied primarily on visual information \cite{Arandjelovic05,Everingham06}, and aimed for the slightly different task of face track labeling; speakers who did not appear in the video frame were not assigned any names, which is common in movies and TV shows. Other available sources of information such as scripts were only used to extract cues about the speakers' names to associate the faces in the videos with their corresponding character name \cite{Everingham06,Tapaswi15,Baeuml13,Sivic09}; however since scripts are not always available, the applicability of these methods is somehow limited. 

Other studies focused on the problem of speaker recognition without naming, using the speech modality as a single source of information. While some of these studies attempted to incorporate the visual modality, their goal was to cluster the speech segments rather than name the speakers \cite{Erzin05,Bost14,Kapsouras15,Bredin16,Hu15,Ren16}. None of these studies used textual information (e.g., dialogue), which prevented them from identifying speaker names.

In our work, we address the task of speaker naming, and propose a new multimodal model that leverages in an unified framework of the visual, speech, and textual modalities that are naturally available while watching a movie. We do not assume the availability of a movie script or a cast list, which makes our model fully unsupervised and easily applicable to unseen movies. 

The paper makes two main contributions. First, we introduce a new unsupervised system for speaker naming for movies and TV shows that exclusively depends on videos and subtitles, and relies on a novel unified optimization framework that fuses visual, textual, and acoustic modalities for speaker naming. Second, we construct and make available a dataset consisting of 24 movies with 31,019 turns manually annotated with character names. Additionally, we also evaluate the role of speaker naming when embedded in an end-to-end memory network model, achieving state-of-the-art performance results on the subtitles task of the MovieQA 2017 Challenge.


\section{Related Work}
\label{sec:related}

The problem of speaker naming in movies has been explored by the computer vision and the speech communities. In the computer vision community, the speaker naming problem is usually considered as a face/person naming problem, in which names are assigned to their corresponding faces on the screen \cite{Everingham06,Cour10,Baeuml13,Haurilet16,Tapaswi15}. On the other hand, the speech community considered the problem as a speaker identification problem, which focuses on recognizing and clustering speakers rather than naming them \cite{Reynolds02,Campbell97}. In this work, we aim to solve the problem of speaker naming in movies, in which we label each segment of the subtitles with its corresponding speaker name whether the speaker's face appeared on in the video or not.

Previous work can be furthered categorized according to the type of supervision used to build the character recognition and speaker recognition models: supervised vs. weakly supervised models. In the movie and television domains, utilizing scripts in addition to subtitles to obtain timestamped speaker information was also studied in \cite{Everingham06,Tapaswi15,Baeuml13,Sivic09}. Moreover, they utilized this information to resolve the ambiguity introduced by co-occurring faces in the same frame. Features were extracted through the period of speaking (detected via lip motion on each face). Then they assigned the face based on candidate names from the time-stamped script. Thus, these studies used speaker recognition as an essential step to construct cast-specific face classifiers. \cite{Tapaswi12} extended the face identification problem to include person tracking. They utilized available face recognition results to learn clothing models for characters to identify person tracks without faces. 

In \cite{Cour10,Haurilet16}, the authors proposed a weakly supervised model depending on subtitles and a character list. They extracted textual cues from the dialog: first, second, and third person references, such as ``I'm Jack'', ``Hey, Jack!'', and ``Jack left''. Using a character list from IMDB, they mapped these references onto true names using minimum edit distance, and then they ascribed the references to face tracks. Other work removed the dependency on a true character list by determining all names through coreference resolution. However, this work also depended on the availability of scripts \cite{Ramanathan14}. In our model, we removed the dependency on both the true cast list and the script, which makes it easier to apply our model to other movies and TV shows. 

Recent work proposed a convolutional neural network (CNN) and Long Short-Term Memory (LSTM) based learning framework to automatically learn a function that combines both facial and acoustic features \cite{Hu15,Ren16}. Using these cues, they tried to learn matching face-audio pairs and non-matching face-audio pairs. They then trained a SVM classifier on the audio-video pairings to discriminate between the non-overlapping speakers. In order to train their models, they manually identified the leading characters in two TV shows, Friends and The Big Bang Theory (BBT), and collected their face tracks and corresponding audio segments using pre-annotated subtitles. Despite the very high performance reported in these studies, it is very hard to generalize their approach since it requires a lot of training data. 

On the other hand, talking faces have been used to improve speaker recognition and diarization in TV shows \cite{Bredin16,Bost14,Li04}. In the case of \cite{Liu08}, they modeled the problem of speaker naming as facial recognition to identify speakers in news broadcasts. This work leveraged optical character recognition to read the broadcasters' names that were displayed on screen, requiring the faces to already be annotated.


\section{Datasets}
\label{sec:data}

Our dataset consists of a mix of TV show episodes and full movies. For the TV show, we use six full episodes of season one of the BBT. The number of named characters in the BBT episodes varies between 5 to 8 characters per episode, and the background noise level is low. Additionally, we also acquired a set of eighteen full movies from different genres, to evaluate how our model works under different conditions. In this latter dataset, the number of named characters ranges between 6 and 37, and it has varied levels of background noise. 

We manually annotated this dataset with the character name of each subtitle segment. To facilitate the annotation process, we built an interface that parses the movies subtitles files, collects the cast list from IMDB for each movie, and then shows one subtitle segment at a time along with the cast list so that the annotator can choose the correct character. Using this tool, human annotators watched the movies and assigned a speaker name to each subtitle segment. If a character name was not mentioned in the dialogue, the annotators labeled it as ``unknown.'' To evaluate the quality of the annotations, five movies in our dataset were double annotated. The Cohen's Kappa inter-annotator agreement score for these five movies is 0.91, which shows a strong level of agreement. 

To clean the data, we removed empty segments, as well as subtitle description parts written between brackets such as ``[groaning]'' and ``[sniffing]''. We also removed segments with two speakers at the same time. We intentionally avoided using any automatic means to split these segments, to preserve the high-quality of our gold standard.

Table \ref{table:datastats} shows the statistics of the collected data. Overall, the dataset consists of 24 videos with a total duration of 40.28 hours, a net dialogue duration of 21.99 hours, and a total of 31,019 turns spoken by 463 different speakers.  Four of the movies in this dataset are used as a development set to develop supplementary systems and to fine tune our model's parameters; the remaining movies are used for evaluation. 

\begin{table}[h]
\centering
    \scalebox{0.8}{
    \begin{tabular}             { l c c c c c } \hline
                                & Min & Max & Mean & $\sigma$ \\ \hline 
    \# characters/video         & 5   & 37  & 17.8 &  9.55  \\ 
    \# Subtitle turns/video     & 488 & 2212 & 1302.4 & 563.06 \\ 
    \# words/turn               & 1 & 28  &  8.02 & 4.157 \\ 
    subtitles duration (sec)    & 0.342 & 9.59 & 2.54 & 1.02 \\
    \hline
    \end{tabular}
    }
    \vspace{-0.2cm}
\caption{Statistics on the annotated movie dataset.}
    \vspace{-0.2cm}
\label{table:datastats}
\end{table}


\section{Data Processing and Representations}
\label{sec:modalities}

We process the movies by extracting several textual, acoustic, and visual features. 

\subsection{Textual Features}

We use the following representations for the textual content of the subtitles:

\noindent \textbf{SkipThoughts} uses a Recurrent Neural Network to capture the underlying semantic and syntactic properties, and map them to a vector representation \cite{Kiros15}. We use their pretrained model to compute a 4,800 dimensional sentence representation for each line in the subtitles.\footnote{https://github.com/ryankiros/skip-thoughts}

\noindent \textbf{TF-IDF} is a traditional weighting scheme in information retrieval. We represent each subtitle as a vector of tf-idf weights, where the length of the vector (i.e., vocabulary size) and the idf scores are obtained from the movie including the subtitle.

\subsection{Acoustic Features}

For each movie in the dataset, we extract the audio from the center channel. The center channel is usually dedicated to the dialogue in movies, while the other audio channels carry the surrounding sounds from the environment and the musical background. Although doing this does not fully eliminate the noise in the audio signal, it still improves the speech-to-noise ratio of the signal. When a movie has stereo sound (left and right channels only), we down-mix both channels of the stereo stream into a mono channel. 

In this work, we use the subtitles timestamps as an estimate of the boundaries that correspond to the uttered speech segments. Usually, each subtitle corresponds to a segment being said by a single speaker. 
We use the subtitle timestamps for segmentation so that we can avoid automatic speaker diarization errors and focus on the speaker naming problem.

To represent the relevant acoustic information from each spoken segment, we use iVectors, which is the state-of-the-art unsupervised approach in speaker verification \cite{Dehak11}. While other deep learning-based speaker embeddings models also exist, we do not have access to enough supervised data to build such models. We train unsupervised iVectors for each movie in the dataset, using the iVector extractor used in \cite{Khorram16}. We extract iVectors of size 40 using a  Gaussian Mixture Model-Universal Background Model (GMM-UBM) with 512 components. Each iVector corresponds to a speech segment uttered by a single speaker. We fine tune the size of the iVectors and the number of GMM-UBM components using the development dataset.

\subsection{Visual Features}

We detect faces in the movies every five frames using the recently proposed MTCNN \cite{zhang2016joint} model, which is pretrained for face detection and facial landmark alignment. Based on the results of face detection, we apply the forward and backward tracker with an implementation of the Dlib library \cite{King09,Danelljan14} to extract face tracks from each video clip. We represent a face track using its best face in terms of detection score, and use the activations of the fc7 layer of pretrained VGG-Face \cite{Parkhi15} network as visual features.

We calculate the distance between the upper lip center and the lower lip center based on the 68-point facial landmark detection implemented in the Dlib library \cite{King09,Kazemi14}. This distance is normalized by the height of face bounding boxes and concatenated across frames to represent the amount of mouth opening. A human usually speaks with lips moving with a certain frequency (3.75 Hz to 7.5 Hz used in this work) \cite{Tapaswi15}. We apply a band-pass filter to amplify the signal of true lip motion in these segments. The overall sum of lip motion is used as the score for the talking face.


\section{Unified Optimization Framework}
\label{sec:approach}

We tackle the problem of speaker naming as a transductive learning problem with constraints. In this approach, we want to use the sparse positive labels extracted from the dialogue and the underlying topological structure of the rest of the unlabeled data. We also incorporate multiple cues extracted from both textual and multimedia information. A unified learning framework is proposed to enable the joint optimization over the automatically labeled and unlabeled data, along with multiple semantic cues.

\subsection{Character Identification and Extraction}
\label{subsec:char}

In this work, we do not consider the set of character names as given because we want to build a model that can be generalized to unseen movies. This strict setting adds to the problem's complexity. To extract the list of characters from the subtitles, we use the Named Entity Recognizer (NER) in the Stanford CoreNLP toolkit \cite{Manning14}. The output is a long list of person names that are mentioned in the dialogue. This list is prone to errors including, but not limited to, nouns that are misclassified by the NER as person's name such as ``Dad" and ``Aye", names that are irrelevant to the movie such as ``Superman" or named animals, or uncaptured character names.  

To clean the extracted names list of each movie, we cluster these names based on string minimum edit distance and their gender. From each cluster, we then pick a name to represent it based on its frequency in the dialogue. The result of this  step consists of name clusters along with their distribution in the dialogue. The distribution of each cluster is the sum of all the counts of its members. To filter out irrelevant characters, we run a name reference classifier, which classifies each name into first, second or third person references. If a name was only mentioned as a third person throughout the whole movie, we discard it from the list of characters. We remove any name cluster that has a total count less than three, which takes care of the misclassified names' reference types. 

\subsection{Grammatical Cues}
\label{subsec:ref}
We use the subtitles to extract the name mentions in the dialogue. These mentions allow us to obtain cues about the speaker name and the absence or the presence of the mentioned character in the surrounding subtitles. Thus, they affect the probability that the mentioned character is the speaker or not. We follow the same name reference categories used in \cite{Cour10,Haurilet16}. We classify a name mention into: first (e.g., ``I'm Sheldon''), second (e.g., ``Oh, hi, Penny'') or third person reference (e.g., ``So how did it go with Leslie?''). The first person reference represents a positive constraint that allows us to label the corresponding iVector of the speaker and his face if it exists during the segment duration. The second person reference represents a multi-instance constraint that suggests that the mentioned name is one of the characters that are present in the scene, which increases the probability of this character to be one of the speakers of the surrounding segments. On the other hand, the third person reference represents a negative constraint, as it suggests that the speaker does not exist in the scene, which lowers the character probability of the character being one of the speakers of the next or the previous subtitle segments. 

To identify first, second and third person references, we train a linear support vector classifier. The first person, the second and third person classifier's training data are extracted and labeled from our development dataset, and fine tuned using 10-fold cross-validation. Table \ref{table:ref_class} shows the results of the classifier on the test data. The average number of first, second and third-person references in each movie are 14.63, 117.21, and 95.71, respectively.

\begin{table}[ht]
\centering
    \scalebox{0.9}{
    \begin{tabular}{l c c c}
    \hline
                        & Precision & Recall & F1-Score \\ \hline
        First Person    & 0.625     & 0.448 & 0.522 \\ 
        Second Person   & 0.844     & 0.863 & 0.853 \\ 
        Third Person    & 0.806     & 0.806 & 0.806 \\ 
        Average / Total & 0.819     & 0.822 & 0.820 \\ \hline
    \end{tabular}
    }
    \vspace{-0.2cm}
\caption{Performance metrics of the reference classifier on the test data.}
\vspace{-0.2cm}
\label{table:ref_class}
\end{table}

\subsection{Unified Optimization Framework}
\label{subsec:semi}
Given a set of data points that consist of $l$  labeled\footnote{Note that in our setup, all the labeled instances are obtained automatically, as described above.} and $u$ unlabeled instances, we apply an  optimization framework to infer the best prediction of speaker names. Suppose we have $l+u$ instances $X=\{x_1,x_2,...,x_l,x_{l+1},...,x_{l+u}\}$ and $K$ possible character names. We also get the dialogue-based positive labels $y_i$ for instances $x_i$, where $y_i$ is a $k$-dimension one-hot vector and $y_{i}^j=1$ if $x_i$ belongs to the class $j$, for every $1\leq i\leq l$ and $1\leq j \leq K$. To name each instance $x_i$, we want to predict another one-hot vector of naming scores $f(x_i)$ for each $x_i$, such that $\mbox{argmax}_{j} f^j(x_i) = z_i$ where $z_i$ is the ground truth number of class for instance $x_i$.

To combine the positive labels and unlabeled data, we define the objective function for predictions $f$ as follows:
\begin{equation}
    \begin{aligned}
    L_{initial}(f) &= \frac{1}{l}\sum_{i=1}^l ||f(x_i)-y_i||^2 \\
    &+ \frac{1}{l+u} \sum_{i=1}^{l+u}\sum_{j=1}^{l+u} w_{ij} ||f(x_i)-f(x_j)|| ^2
    \end{aligned}
    \label{eqn:semi}
\end{equation}

Here $w_{ij}$ is the similarity between $x_i$ and $x_j$, which is calculated as the weighted sum of textual, acoustic and visual similarities. The inverse Euclidean distance is used as similarity function for each modality. The weights for different modalities are selected as hyperparameters and tuned on the development set. This objective leads to a convex loss function which is easier to optimize over feasible predictions.

Besides the positive labels obtained from first person name references, we also introduce other semantic constraints and cues to enhance the power of our proposed approach. We implement the following four types of constraints:

\mysubsection{Multiple Instance Constraint.} Although the second person references cannot directly provide positive constraints, they imply that the mentioned characters have high probabilities to be in this conversation. Following previous work \cite{Cour10}, we incorporate the second person references as multiple instances constraints into our optimization: if $x_i$ has a second person reference $j$, we encourage $j$ to be assigned to its neighbors, i.e., its adjacent subtitles with similar timestamps. For the implementation, we simply include multiple instances constraints as a variant of positive labels with decreasing weights $s$, where $s=1/(l-i)$ for each neighbor $x_l$.

\mysubsection{Negative Constraint.} For the third person references, the mentioned characters may not occur in the conversation and movies. So we treat them as negative constraints, which means they imply that the mentioned characters should not be assigned to corresponding instances. This constraint is formulated as follows:
\vspace{-0.2cm}
\begin{equation}
    L_{neg}(f) = \sum_{(i,j)\in N} [f^j(x_i)]^2
\end{equation}
where $N$ is the set of negative constraints $x_i$ doesn't belong class $j$.

\mysubsection{Gender Constraint.} We train a voice-based gender classifier by using the subtitles segments from the four movies in our development dataset (5,543 segments of subtitles). We use the segments in which we know the speaker's name and manually obtain the ground truth gender label from IMDB. We extract the signal energy, 20 Mel-frequency cepstral coefficients (MFCCs) along with their first and second derivatives, in addition to time- and frequency-based absolute fundamental frequency (f0) statistics as features to represent each segment in the subtitles. The f0 statistics has been found to improve the automatic gender detection performance for short speech segments \cite{Levitan16}, which fits our case since the median duration of the dialogue turns in our dataset is 2.6 seconds.

The MFCC features are extracted using a step size of 16 msec over a 64 msec window using the method from  \cite{Mathieu10}, while the f0 statistics are extracted using a step size of 25 msec over a 50 msec window as the default configuration in \cite{Eyben13}. We then use these features to train a logistic regression classifier using the Scikit-learn library \cite{scikit-learn}. The average accuracy of the gender classifier on a 10-fold cross-validation is 0.8867. 

Given the results for the gender classification of audio segments and character names, we define the gender loss to penalize inconsistency between the predicted gender and character names:
\vspace{-0.1cm}
\begin{equation}
    \begin{aligned}
    L_{gender}(f) &= \sum_{(i,j)\in Q_1}P_{ga}(x_i) (1-P_{gn}(j))f^j(x_i)\\
    &+\sum_{(i,j)\in Q_2}(1-P_{ga}(x_i))P_{gn}(j)f^j(x_i)
    \end{aligned}
\end{equation}
where $P_{ga(x_i)}$ is the probability for instance $x_i$ to be a male, and $P_{gn(j)}$ is the probability for name $j$ to be a male, and $Q_1=\{(i,j)|P_{ga}(x_i)<0.5,P_{gn}(j)>0.5\}$, $Q_2=\{(i,j)|P_{ga}(x_i)>0.5,P_{gn}(j)<0.5\}$.

\mysubsection{Distribution Constraint.} We automatically analyze the dialogue and extract the number of mentions of each character in the subtitles using Stanford CoreNLP and string matching to capture names that are missed by the named entity recognizer. We then filter the resulting counts by removing third person mention references of each name as we assume that this character does not appear in the surrounding frames. We use the results to estimate the distribution of the speaking characters and their importance in the movies. The main goal of this step is to construct a prior probability distribution for the speakers in each movie.

To encourage our predictions to be consistent with the dialogue-based priors, we penalize the square error between the distributions of predictions and name mentions priors in the following equation:
\vspace{-0.5cm}
\begin{equation}
    L_{dis}(f) = \sum_{j=1}^K (\sum(f^j(x_i)) - d_j)^2
\end{equation}

\noindent where $d_j$ is the ratio of name $j$ mentions in all subtitles.

\mysubsection{Final Framework.} Combining the loss in Eqn. \ref{eqn:semi} and multiple losses with different constraints, we obtain our unified optimization problem:
\vspace{-0.02cm}
\begin{equation}
\begin{aligned}
f^{\ast} &= \arg\min_{f} \lambda_1 L_{initial}(f)+ \lambda_2 L_{MI}(f) \\
& + \lambda_3 L_{neg}(f) +  \lambda_4 L_{gender}(f) + \lambda_5 L_{dis}(f)
\end{aligned}
\label{eqn:unify}
\end{equation}

All of the $\lambda$s are hyper-parameters to be tuned on development set. We also include the constraint that predictions for different character names must sum to 1. We solve this constrained optimization problem with projected gradient descent (PGD). Our optimization problem in Eqn. \ref{eqn:unify} is guaranteed to be a convex optimization problem and therefore projected gradient descent is guaranteed to stop with global optima. PGD usually converges after 800 iterations.

\section{Evaluation}
\label{sec:eval}

We model our task as a classification problem, and use the unified optimization framework described earlier to assign a character name to each subtitle. 

Since our dataset is highly unbalanced, with a few main characters usually dominating the entire dataset, we adopt the weighted F-score as our evaluation metric, instead of using an accuracy metric or a micro-average F-score. This allows us to take into account that most of the characters have only a  few spoken subtitle segments, while at the same time placing emphasis on the main characters. This leads sometimes to an average weighted F-score that is not between the average precision and recall.

\begin{table}[htb]
\centering
    \scalebox{0.75}{
    \begin{tabular}{ l c c c } \hline
               & Precision &  Recall &  F-score   \\\hline
    B1: MFMC                            & 0.0910 & 0.2749 & 0.1351     \\
    B2: DRA                             & 0.2256 & 0.1819 & 0.1861     \\ 
    B3: Gender-based DRA                & 0.2876 & 0.2349 & 0.2317     \\ \hline
    Our Model (Skip-thoughts)\mbox{*}   & 0.3468 & 0.2869 & 0.2680     \\
    Our Model (TF-IDF)\mbox{*}          & 0.3579 & 0.2933 & 0.2805     \\
    Our Model (iVectors)                & 0.2151 & 0.2347 & 0.1786     \\
    Our Model (Visual)\mbox{*}          & 0.3348 & 0.2659 & 0.2555     \\
    Our Model (Visual+iVectors)\mbox{*} & 0.3371 & 0.2720 & 0.2617     \\
    Our Model (TF-IDF+iVectors)\mbox{*} & 0.3549 & 0.2835 & 0.2643     \\
    Our Model (TF-IDF+Visual)\mbox{*}   & 0.3385 & 0.2975 & 0.2821     \\ \hline
    Our Model (all)\mbox{*}    & \textbf{0.3720} & \textbf{0.3108} & \textbf{0.2920} \\ \hline
    \end{tabular}
    }
    \vspace{-0.2cm}
\caption{Comparison between the average of macro-weighted average of precision, recall and f-score of the baselines and our model. * means statistically significant (t-test p-value $<$ 0.05) when compared to baseline B3.}
\vspace{-0.2cm}
\label{table:eval}
\end{table}

One aspect that is important to note is that characters are often referred to using different names. For example, in the movie ``The Devil's Advocate," the character Kevin Lomax is also referred to as Kevin or Kev. In more complicated situations, characters may even have multiple identities, such as the character Saul Bloom in the movie ``Ocean's Eleven,'' who pretends to be another character named Lyman Zerga. Since our goal is to assign names to speakers, and not necessarily solve this coreference problem, we consider the assignment of the subtitle segments to any of the speaker's aliases to be correct. Thus, during the evaluation, we map all the characters' aliases from our model's output to the names in the ground truth annotations. Our mapping does not include other referent nouns such as ``Dad,'' ``Buddy,'' etc.; if a segment gets assigned to any such terms, it is considered a misprediction.

We compare our model against three baselines:

\noindent \textbf{B1: Most-frequently mentioned character} consists of selecting the most frequently mentioned character in the dialogue as the speaker for all the subtitles. Even though it is a simple baseline, it achieves an accuracy of 27.1\%, since the leading characters tend to speak the most in the movies.

\noindent \textbf{B2: Distribution-driven random assignment} consists of randomly assigning character names according to a distribution that reflects their fraction of mentions in all the subtitles. 

\noindent \textbf{B3: Gender-based distribution-driven random assignment} consists of selecting the speaker names based on the voice-based gender detection classifier. This baseline randomly selects the character name that matches the speaker's gender according to the distribution of mentions of the names in the matching gender category. 

The results obtained with our proposed unified optimization framework and the three baselines are shown in Table \ref{table:eval}. We also report the performance of the optimization framework using different combinations of the three modalities. The model that uses all three modalities achieves the best results, and outperforms the strongest baseline (B3) by more than 6\% absolute in average weighted F-score. It also significantly outperforms the usage of the visual and acoustic features combined, which have been frequently used together in previous work, suggesting the importance of textual features in this setting.

\begin{figure}[!tb]
\begin{center}
    \vspace{-0.09cm}
   \begin{subfigure}[t]{1\linewidth}
        \centering
        \includegraphics[width=\linewidth]{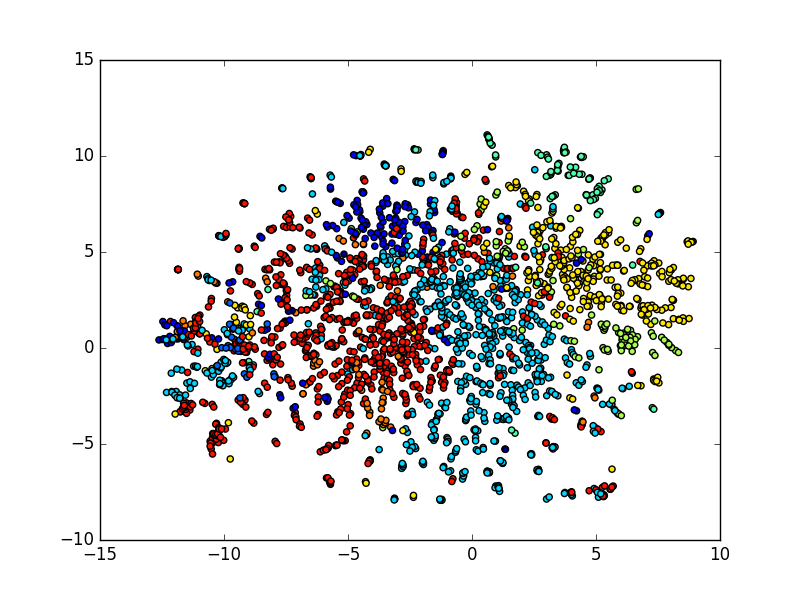}
        \vspace{-0.5cm}
        \caption{The Big Bang Theory}
    \end{subfigure}%
    \vspace{-0.1cm}
    \begin{subfigure}[t]{1\linewidth}
        \centering
        \includegraphics[width=\linewidth]{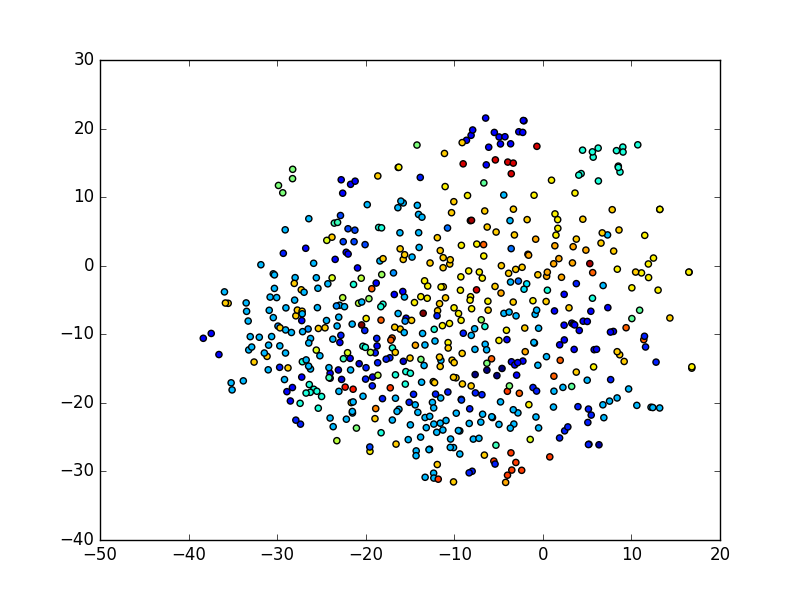}
        \vspace{-0.5cm}
        \caption{Titanic}
    \end{subfigure}%
\end{center}
   \vspace{-0.2cm}
   \caption{For each speech segment, we applied t-SNE \cite{van2014accelerating} on their corresponding iVectors. The points with the same color represent instances with the same character name.}
   \vspace{-0.3cm}
\label{fig:ivec_cluster}
\end{figure}

The ineffectiveness of the iVectors might be a result of the background noise and music, which are difficult to remove from the speech signal. Figure \ref{fig:ivec_cluster} shows the t-Distributed Stochastic Neighbor Embedding (t-SNE) \cite{van2014accelerating}, which is a nonlinear dimensionality reduction technique that models points in such a way that similar vectors are modeled by nearby points and dissimilar objects are modeled by distant points, visualization of the iVectors over the whole BBT show and the movie ``Titanic.'' In the BBT there is almost no musical background or background noise, while, Titanic has musical background in addition to the background noise such as the screams of the drowning people. From the graph, the difference between the quality of the iVectors clusters on different noise-levels is clear. 

Table \ref{table:eval_loss} shows the effect of adding components of our loss function to the initial loss $L_{init}$ function. The performance of the model using only $L_{init}$  without the other parts is very low due to the sparsity of first person references and errors that the person reference classifier introduces.

\begin{table}[htb]
\centering
    \scalebox{0.8}{
    \begin{tabular}{ l c c c } \hline
                                           & Precision & Recall & F-score \\ \hline
    $L_{initial}$                          & 0.0631 &  0.1576 &  0.0775   \\
    $L_{initial}$ + $L_{gender}$           & 0.1160 &  0.1845 &  0.1210   \\
    $L_{initial}$ + $L_{negative}$         & 0.0825 &  0.0746 &  0.0361   \\
    $L_{initial}$ + $L_{distribution}$     & 0.1050 &  0.1570 &  0.0608   \\
    $L_{initial}$ + $L_{Multiple Instance}$& 0.3058 &  0.2941 &  0.2189   \\
    \hline
    \end{tabular}
    }
    \vspace{-0.2cm}
\caption{Analysis of the effect of adding each component of the loss function to the initial loss.}
\vspace{-0.2cm}
\label{table:eval_loss}
\end{table}

In order to analyze the effect of the errors that several of the modules (e.g., gender and name reference classifiers) propagate into the system, we also test our framework by replacing each one of the components with its ground truth information. As seen in Table \ref{tab:evalgt}, the results obtained in this setting show significant improvement with the replacement of each component in our framework, which suggests that additional work on these components will have positive implications on the overall system. 
\begin{table}[htb]
\centering
    \scalebox{0.8}{
    \begin{tabular}{ l c c c } \hline
                            & Precision & Recall & F-score \\ \hline
    Our Model               & 0.3720 & 0.3108 & 0.2920    \\
    Voice Gender (VG)       & 0.4218 & 0.3449 & 0.3259     \\
    VG + Name Gender (NG)   & 0.4412 & 0.3790 & 0.3645     \\
    VG + NG + Name Ref      & 0.4403 & 0.3938 & 0.3748     \\
    
    \hline
    \end{tabular}
    }
    \vspace{-0.1cm}
\caption{Comparison between our model while replacing different components with their ground truth information.}
\vspace{-0.1cm}
\label{tab:evalgt}
\end{table}

\section{Speaker Naming for Movie Understanding}

Identifying speakers is a critical task for understanding the dialogue and storyline in movies. MovieQA is a challenging dataset for movie understanding. 
The dataset consists of 14,944 multiple choice questions about 408 movies. Each question has five answers and only one of them is correct. The dataset is divided into three splits: train, validation, and test according to the movie titles. Importantly, there are no overlapping movies between the splits. Table \ref{table:MQAExamples} shows examples of the question and answers in the MovieQA dataset.

\begin{table*}[ht]
\centering
\scalebox{0.87}{ 
\begin{tabular}{l l p{4.8cm}p{4cm}}
\hline
Movie & Question & \multicolumn{2}{l}{Answers}\\\hline

\multirow{3}{1.2cm}{Fargo}& 
\multirow{3}{5cm}{What did Mike's wife, as he says, die from?} & 
    A1: She was killed &\textbf{A2: Breast cancer} \\
    &    & A3: Leukemia &  A4: Heart disease \\
    &    &
    \multicolumn{2}{l}{A5: Complications due to child birth}
\\\hline

\multirow{3}{1.2cm}{Titanic}& 
\multirow{3}{5cm}{What does Rose ask Jack to do in her room?}
&
    A1: Sketch her in her best dress & A2: Sketch her nude  \\
    &    & A3: Take a picture of her nude & \textbf{A4: Paint her nude} \\
    &    &
    \multicolumn{2}{l}{A5: Take a picture of her in her best dress}
        \\\hline
\end{tabular}
}
\vspace{-0.2cm}
\caption{  
Example of questions and answers from the MQA benchmark. The answers in bold are the correct answers to their corresponding question.}
\vspace{-0.2cm}
\label{table:MQAExamples}
\end{table*}

The MovieQA 2017 Challenge\footnote{http://movieqa.cs.toronto.edu/workshops/iccv2017/} consists of six different tasks according to the source of information used to answer the questions. Given that for many of the movies in the dataset the videos are not completely available, we develop our initial system so that it only relies on the subtitles; we thus participate in the challenge subtitles task, which includes the dialogue (without the speaker information) as the only source of information to answer questions.

\begin{figure}[t]
\begin{center}
 \scalebox{0.99}{
   \includegraphics[width=\linewidth]{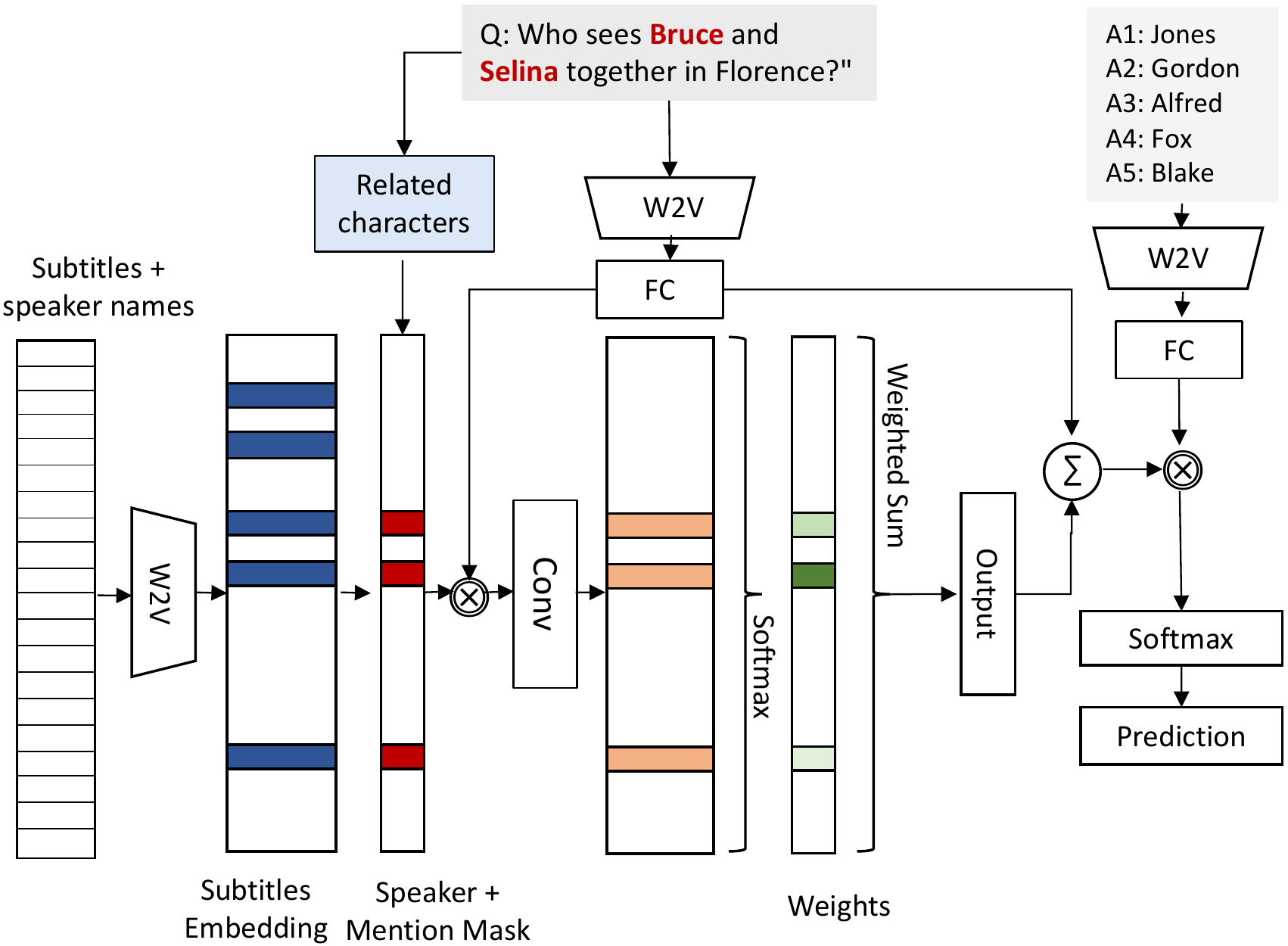}
   }
\end{center}
\vspace{-0.2cm}
   \caption{The diagram describing our Speaker-based Convolutional Memory Network (SC-MemN2N) model.}
\vspace{-0.2cm}
\label{fig:scmemn}
\end{figure}

To demonstrate the effectiveness of our speaker naming approach, we design a model based on an  end-to-end memory network \cite{Sukhbaatar15}, namely Speaker-based Convolutional Memory Network (SC-MemN2N), which relies on the MovieQA dataset, and integrates the speaker naming approach as a component in the network. Specifically, we use our speaker naming framework to infer the name of the speaker for each segment of the subtitles, and prepend the predicted speaker name to each turn in the subtitles.\footnote{We strictly follow the challenge rules, and only use text to infer the speaker names.} To represent the movie subtitles, we represent each turn in the subtitles as the mean-pooling of a 300-dimension pretrained word2vec \cite{Mikolov13} representation of each word in the sentence. We similarly represent the input questions and their corresponding answers. Given a question, we use the SC-MemN2N memory to find an answer. For questions asking about specific characters, we keep the memory slots that have the characters in question as speakers or mentioned in, and mask out the rest of the memory slots. Figure \ref{fig:scmemn} shows the architecture of our model.

\begin{figure}[htb]
\begin{center}
 \scalebox{0.99}{
  \includegraphics[width=\linewidth, height=5cm]{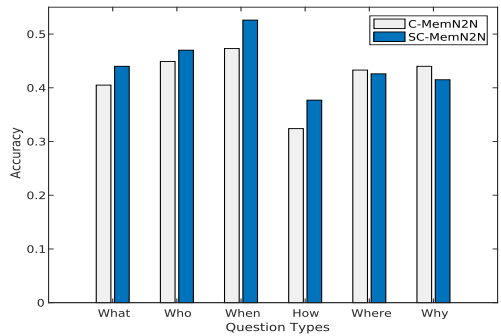}
  }
\end{center}
\vspace{-0.2cm}
  \caption{Accuracy comparison according to question type.}
\vspace{-0.2cm}
\label{fig:scmemn_q}
\end{figure}

Table \ref{table:movieqa_results} includes the results of our system on the validation and test sets, along with the best systems introduced in previous work, showing that our SC-MemN2N achieves the best performance. Furthermore, to measure the effectiveness of adding the speaker names and masking, we test our model after removing the names from the network (C-MemN2N). As seen from the results, the gain of SC-MemN2N is statistically significant\footnote{Using a t-test p-value$<$0.05 with 1,000 folds each containing 20 samples.} compared to a version of the system that does not include the speaker names (C-MemN2N). Figure \ref{fig:scmemn_q} shows the performance of both C-MemN2N and SC-MemN2N models by question type. The results suggest that our speaker naming helps the model better distinguish between characters, and that prepending the speaker names to the subtitle segments improves the ability of the memory network to correctly identify the supporting facts from the story that answers a given question.

\begin{table}[!htb]
\centering
    \scalebox{0.9}{
        \begin{tabular}{ l  c c} \hline
        \multirow{2}{0em}{Method}       & \multicolumn{2}{c}{Subtitles} \\ 
                                        & val           & test          \\ \hline
        SSCB-W2V \cite{Tapaswi16}       & 24.8          & 23.7          \\
        SSCB-TF-IDF \cite{Tapaswi16}    & 27.6          & 26.5          \\
        SSCB Fusion \cite{Tapaswi16}    & 27.7          & -             \\
        MemN2N \cite{Tapaswi16}         & 38.0          & 36.9          \\
        Understanding visual regions    & -             & 37.4          \\
        RWMN   \cite{Na17}              & 40.4          & 38.5          \\ \hline
        C-MemN2N (w/o SN)               & 40.6          & -             \\
        SC-MemN2N (Ours)                & \textbf{42.7} & \textbf{39.4} \\
        \hline
        \end{tabular}
    }
    \vspace{-0.1cm}
\caption{Performance comparison for the subtitles task on the MovieQA 2017 Challenge on both validation and test sets. We compare our models with the best existing models (from the challenge leaderboard). 
}
\vspace{-0.1cm}
\label{table:movieqa_results}
\end{table}

\section{Conclusion}
\label{sec:conclusion}

In this paper, we proposed a unified optimization framework for the task of speaker naming in movies. We addressed this task under a difficult setup, without a cast-list, without supervision from a script, and dealing with the complicated conditions of real movies. Our model includes textual, visual, and acoustic modalities, and incorporates several grammatical and acoustic constraints. Empirical experiments on a movie dataset demonstrated the effectiveness of our proposed method with respect to several competitive baselines. We also showed that an SC-MemN2N model that leverages our speaker naming model can achieve state-of-the-art results on the subtitles task of the MovieQA 2017 Challenge.


The dataset annotated with character names introduced in this paper is publicly available from \url{http://lit.eecs.umich.edu/downloads.html}.

\section*{Acknowledgments}
We would like to thank the anonymous reviewers for their valuable comments and suggestions. This work is supported by a Samsung research grant and by a DARPA grant HR001117S0026-AIDA-FP-045. 

\bibliography{naaclhlt2018}

\begin{thebibliography}{}
\expandafter\ifx\csname natexlab\endcsname\relax\def\natexlab#1{#1}\fi

\bibitem[{Arandjelovic and Zisserman(2005)}]{Arandjelovic05}
Ognjen Arandjelovic and Andrew Zisserman. 2005.
\newblock Automatic face recognition for film character retrieval in
  feature-length films.
\newblock In {\em Computer Vision and Pattern Recognition (CVPR)\/}.

\bibitem[{B\"{a}uml et~al.(2013)B\"{a}uml, Tapaswi, and
  Stiefelhagen}]{Baeuml13}
Martin B\"{a}uml, Makarand Tapaswi, and Rainer Stiefelhagen. 2013.
\newblock Semi-supervised learning with constraints for person identification
  in multimedia data.
\newblock In {\em IEEE Conference on Computer Vision and Pattern Recognition
  (CVPR)\/}.

\bibitem[{Bost and Linares(2014)}]{Bost14}
Xavier Bost and Georges Linares. 2014.
\newblock Constrained speaker diarization of tv series based on visual
  patterns.
\newblock In {\em IEEE Spoken Language Technology Workshop (SLT)\/}.

\bibitem[{Bredin and Gelly(2016)}]{Bredin16}
Herv{\'e} Bredin and Gr{\'e}gory Gelly. 2016.
\newblock Improving speaker diarization of tv series using talking-face
  detection and clustering.
\newblock In {\em Proceedings of the 24th Annual ACM Conference on
  Multimedia\/}.

\bibitem[{Campbell(1997)}]{Campbell97}
Joseph~P Campbell. 1997.
\newblock Speaker recognition: A tutorial.
\newblock {\em Proceedings of the IEEE\/} .

\bibitem[{Cour et~al.(2010)Cour, Sapp, Nagle, and Taskar}]{Cour10}
Timothee Cour, Benjamin Sapp, Akash Nagle, and Ben Taskar. 2010.
\newblock Talking pictures: Temporal grouping and dialog-supervised person
  recognition.
\newblock In {\em IEEE Conference on Computer Vision and Pattern Recognition
  (CVPR)\/}.

\bibitem[{Danelljan et~al.(2014)Danelljan, H{\"a}ger, Khan, and
  Felsberg}]{Danelljan14}
Martin Danelljan, Gustav H{\"a}ger, Fahad Khan, and Michael Felsberg. 2014.
\newblock Accurate scale estimation for robust visual tracking.
\newblock In {\em British Machine Vision Conference (BMVC)\/}.

\bibitem[{Dehak et~al.(2011)Dehak, Kenny, Dehak, Dumouchel, and
  Ouellet}]{Dehak11}
Najim Dehak, Patrick~J Kenny, R{\'e}da Dehak, Pierre Dumouchel, and Pierre
  Ouellet. 2011.
\newblock Front-end factor analysis for speaker verification.
\newblock {\em IEEE Transactions on Audio, Speech, and Language Processing\/} .

\bibitem[{Erzin et~al.(2005)Erzin, Yemez, and Tekalp}]{Erzin05}
Engin Erzin, Y{\"u}cel Yemez, and A~Murat Tekalp. 2005.
\newblock Multimodal speaker identification using an adaptive classifier
  cascade based on modality reliability.
\newblock {\em IEEE Transactions on Multimedia\/} .

\bibitem[{Everingham et~al.(2006)Everingham, Sivic, and
  Zisserman}]{Everingham06}
Mark Everingham, Josef Sivic, and Andrew Zisserman. 2006.
\newblock ``hello! my name is... buffy'' -- automatic naming of characters in
  tv video.
\newblock In {\em BMVC\/}.

\bibitem[{Eyben et~al.(2013)Eyben, Weninger, Gross, and Schuller}]{Eyben13}
Florian Eyben, Felix Weninger, Florian Gross, and Bj\"orn Schuller. 2013.
\newblock Recent developments in {openSMILE}, the munich open-source multimedia
  feature extractor.
\newblock In {\em Proceedings of the 21st ACM International Conference on
  Multimedia\/}.

\bibitem[{Haurilet et~al.(2016)Haurilet, Tapaswi, Al-Halah, and
  Stiefelhagen}]{Haurilet16}
Monica-Laura Haurilet, Makarand Tapaswi, Ziad Al-Halah, and Rainer
  Stiefelhagen. 2016.
\newblock Naming tv characters by watching and analyzing dialogs.
\newblock In {\em IEEE Winter Conference on Applications of Computer Vision
  (WACV)\/}.

\bibitem[{Hu et~al.(2015)Hu, Ren, Dai, Yuan, Xu, and Wang}]{Hu15}
Yongtao Hu, Jimmy~SJ Ren, Jingwen Dai, Chang Yuan, Li~Xu, and Wenping Wang.
  2015.
\newblock Deep multimodal speaker naming.
\newblock In {\em Proceedings of the 23rd Annual ACM Conference on Multimedia
  Conference\/}.

\bibitem[{Kapsouras et~al.(2015)Kapsouras, Tefas, Nikolaidis, and
  Pitas}]{Kapsouras15}
Ioannis Kapsouras, Anastasios Tefas, Nikos Nikolaidis, and Ioannis Pitas. 2015.
\newblock Multimodal speaker diarization utilizing face clustering information.
\newblock In {\em International Conference on Image and Graphics\/}.

\bibitem[{Kazemi and Sullivan(2014)}]{Kazemi14}
Vahid Kazemi and Josephine Sullivan. 2014.
\newblock One millisecond face alignment with an ensemble of regression trees.
\newblock In {\em IEEE Conference on Computer Vision and Pattern Recognition
  (CVPR)\/}.

\bibitem[{Khorram et~al.(2016)Khorram, Gideon, McInnis, and
  Provost}]{Khorram16}
Soheil Khorram, John Gideon, Melvin McInnis, and Emily~Mower Provost. 2016.
\newblock Recognition of depression in bipolar disorder: Leveraging cohort and
  person-specific knowledge.
\newblock In {\em Interspeech\/}.

\bibitem[{King(2009)}]{King09}
Davis~E King. 2009.
\newblock Dlib-ml: A machine learning toolkit.
\newblock {\em Journal of Machine Learning Research\/} .

\bibitem[{Kiros et~al.(2015)Kiros, Zhu, Salakhutdinov, Zemel, Urtasun,
  Torralba, and Fidler}]{Kiros15}
Ryan Kiros, Yukun Zhu, Ruslan~R Salakhutdinov, Richard Zemel, Raquel Urtasun,
  Antonio Torralba, and Sanja Fidler. 2015.
\newblock Skip-thought vectors.
\newblock In {\em Advances in neural information processing systems\/}.

\bibitem[{Levitan et~al.(2016)Levitan, Mishra, and Bangalore}]{Levitan16}
Sarah~Ita Levitan, Taniya Mishra, and Srinivas Bangalore. 2016.
\newblock Automatic identification of gender from speech.
\newblock {\em Speech Prosody\/} .

\bibitem[{Li et~al.(2004)Li, Narayanan, and Kuo}]{Li04}
Ying Li, Shrikanth~S Narayanan, and C-C~Jay Kuo. 2004.
\newblock Adaptive speaker identification with audiovisual cues for movie
  content analysis.
\newblock {\em Pattern Recognition Letters\/} .

\bibitem[{Liu et~al.(2008)Liu, Jiang, and Huang}]{Liu08}
Chunxi Liu, Shuqiang Jiang, and Qingming Huang. 2008.
\newblock Naming faces in broadcast news video by image google.
\newblock In {\em Proceedings of the 16th ACM international conference on
  Multimedia\/}.

\bibitem[{Manning et~al.(2014)Manning, Surdeanu, Bauer, Finkel, Bethard, and
  McClosky}]{Manning14}
Christopher~D. Manning, Mihai Surdeanu, John Bauer, Jenny Finkel, Steven~J.
  Bethard, and David McClosky. 2014.
\newblock The {Stanford} {CoreNLP} natural language processing toolkit.
\newblock In {\em Association for Computational Linguistics (ACL) System
  Demonstrations\/}.

\bibitem[{Mathieu et~al.(2010)Mathieu, Essid, Fillon, Prado, and
  Richard}]{Mathieu10}
Benoit Mathieu, Slim Essid, Thomas Fillon, Jacques Prado, and Gaël Richard.
  2010.
\newblock Yaafe, an easy to use and efficient audio feature extraction
  software.
\newblock In {\em Proceedings of the 11th International Society for Music
  Information Retrieval Conference\/}.

\bibitem[{Mikolov et~al.(2013)Mikolov, Sutskever, Chen, Corrado, and
  Dean}]{Mikolov13}
Tomas Mikolov, Ilya Sutskever, Kai Chen, Greg~S Corrado, and Jeff Dean. 2013.
\newblock Distributed representations of words and phrases and their
  compositionality.
\newblock In {\em Advances in Neural Information Processing Systems (NIPS)\/}.

\bibitem[{Na et~al.(2017)Na, Lee, Kim, and Kim}]{Na17}
Seil Na, Sangho Lee, Jisung Kim, and Gunhee Kim. 2017.
\newblock A read-write memory network for movie story understanding.
\newblock In {\em International Conference on Computer Vision (ICCV)\/}.

\bibitem[{Parkhi et~al.(2015)Parkhi, Vedaldi, and Zisserman}]{Parkhi15}
O.~M. Parkhi, A.~Vedaldi, and A.~Zisserman. 2015.
\newblock Deep face recognition.
\newblock In {\em British Machine Vision Conference (BMVC)\/}.

\bibitem[{Pedregosa et~al.(2011)Pedregosa, Varoquaux, Gramfort, Michel,
  Thirion, Grisel, Blondel, Prettenhofer, Weiss, Dubourg, Vanderplas, Passos,
  Cournapeau, Brucher, Perrot, and Duchesnay}]{scikit-learn}
F.~Pedregosa, G.~Varoquaux, A.~Gramfort, V.~Michel, B.~Thirion, O.~Grisel,
  M.~Blondel, P.~Prettenhofer, R.~Weiss, V.~Dubourg, J.~Vanderplas, A.~Passos,
  D.~Cournapeau, M.~Brucher, M.~Perrot, and E.~Duchesnay. 2011.
\newblock Scikit-learn: Machine learning in {P}ython.
\newblock {\em Journal of Machine Learning Research\/} .

\bibitem[{Ramanathan et~al.(2014)Ramanathan, Joulin, Liang, and
  Fei-Fei}]{Ramanathan14}
V.~Ramanathan, A.~Joulin, P.~Liang, and L.~Fei-Fei. 2014.
\newblock Linking people with ``their'' names using coreference resolution.
\newblock In {\em IEEE Conference on European Conference on Computer Vision
  (ECCV)\/}.

\bibitem[{Ren et~al.(2016)Ren, Hu, Tai, Wang, Xu, Sun, and Yan}]{Ren16}
Jimmy Ren, Yongtao Hu, Yu-Wing Tai, Chuan Wang, Li~Xu, Wenxiu Sun, and Qiong
  Yan. 2016.
\newblock Look, listen and learn—a multimodal lstm for speaker
  identification.
\newblock In {\em Thirtieth AAAI Conference on Artificial Intelligence\/}.

\bibitem[{Reynolds(2002)}]{Reynolds02}
Douglas~A Reynolds. 2002.
\newblock An overview of automatic speaker recognition technology.
\newblock In {\em Acoustics, speech, and signal processing (ICASSP)\/}.

\bibitem[{Sivic et~al.(2009)Sivic, Everingham, and Zisserman}]{Sivic09}
Josef Sivic, Mark Everingham, and Andrew Zisserman. 2009.
\newblock ``who are you?'' -- learning person specific classifiers from video.
\newblock In {\em IEEE Conference on Computer Vision and Pattern
  Recognition(CVPR)\/}.

\bibitem[{Sukhbaatar et~al.(2015)Sukhbaatar, Weston, Fergus
  et~al.}]{Sukhbaatar15}
Sainbayar Sukhbaatar, Jason Weston, Rob Fergus, et~al. 2015.
\newblock End-to-end memory networks.
\newblock In {\em Advances in neural information processing systems (NIPS)\/}.

\bibitem[{Tapaswi et~al.(2012)Tapaswi, B{\"a}uml, and Stiefelhagen}]{Tapaswi12}
Makarand Tapaswi, Martin B{\"a}uml, and Rainer Stiefelhagen. 2012.
\newblock ``knock! knock! who is it?'' probabilistic person identification in
  tv-series.
\newblock In {\em IEEE Conference on Computer Vision and Pattern Recognition
  (CVPR)\/}.

\bibitem[{Tapaswi et~al.(2014)Tapaswi, B{\"a}uml, and Stiefelhagen}]{Tapaswi14}
Makarand Tapaswi, Martin B{\"a}uml, and Rainer Stiefelhagen. 2014.
\newblock Storygraphs: Visualizing character interactions as a timeline.
\newblock In {\em IEEE Conference on Computer Vision and Pattern Recognition
  (CVPR)\/}.

\bibitem[{Tapaswi et~al.(2015)Tapaswi, B{\"a}uml, and Stiefelhagen}]{Tapaswi15}
Makarand Tapaswi, Martin B{\"a}uml, and Rainer Stiefelhagen. 2015.
\newblock Improved weak labels using contextual cues for person identification
  in videos.
\newblock In {\em IEEE International Conference and Workshops on Automatic Face
  and Gesture Recognition (FG)\/}.

\bibitem[{Tapaswi et~al.(2016)Tapaswi, Zhu, Stiefelhagen, Torralba, Urtasun,
  and Fidler}]{Tapaswi16}
Makarand Tapaswi, Yukun Zhu, Rainer Stiefelhagen, Antonio Torralba, Raquel
  Urtasun, and Sanja Fidler. 2016.
\newblock Movieqa: Understanding stories in movies through question-answering.
\newblock In {\em IEEE Conference on Computer Vision and Pattern Recognition
  (CVPR)\/}.

\bibitem[{Van Der~Maaten(2014)}]{van2014accelerating}
Laurens Van Der~Maaten. 2014.
\newblock Accelerating t-sne using tree-based algorithms.
\newblock {\em Journal of machine learning research\/} .

\bibitem[{Zhang et~al.(2016)Zhang, Zhang, Li, and Qiao}]{zhang2016joint}
Kaipeng Zhang, Zhanpeng Zhang, Zhifeng Li, and Yu~Qiao. 2016.
\newblock Joint face detection and alignment using multitask cascaded
  convolutional networks.
\newblock {\em IEEE Signal Processing Letters\/} .

\bibitem[{Zhang et~al.(2009)Zhang, Xu, Lu, and Huang}]{Zhang09}
Yi-Fan Zhang, Changsheng Xu, Hanquig Lu, and Yeh-Min Huang. 2009.
\newblock Character identification in feature-length films using global
  face-name matching.
\newblock In {\em IEEE Transactions on Multimedia\/}.

\end{thebibliography}
\bibliographystyle{acl_natbib}

\end{document}